\pdfoutput=1

\documentclass[11pt]{article}

\usepackage[final]{acl}

\usepackage{times}
\usepackage{latexsym}
\usepackage{float}

\usepackage[T1]{fontenc}

\usepackage[utf8]{inputenc}

\usepackage{microtype}

\usepackage{inconsolata}

\usepackage{graphicx}

\usepackage{tabularx}
\usepackage{array}
\newcolumntype{Z}{>{\centering\arraybackslash}X}

\title{Large Language Models in Bioinformatics: A Survey}

\author{
\bf Zhenyu Wang$^{\heartsuit}$$^{\spadesuit}$$^{\ast}$,
Zikang Wang$^{\clubsuit}$$^{\heartsuit}$$^{\ast}$,
Jiyue Jiang$^{\heartsuit}$\thanks{Equal Contribution}$^{\dagger}$,
Pengan Chen$^{\diamondsuit}$,\\
\bf Xiangyu Shi$^{\heartsuit}$, 
Yu Li$^{\heartsuit}$\thanks{Corresponding Authors}\\
$^{\heartsuit}$ The Chinese University of Hong Kong,
$^{\spadesuit}$ Peking University Third Hospital, \\
$^{\clubsuit}$ The Hong Kong Polytechnic University,
$^{\diamondsuit}$ The University of Hong Kong \\
{\tt
1810301343@bjmu.edu.cn,
zikang.wang@connect.polyu.hk,}\\
{\tt
jiangjy@link.cuhk.edu.hk,
sxysxygm@gmail.com,
liyu@cse.cuhk.edu.hk
}
}

\begin{document}
\maketitle
\begin{abstract}
Large Language Models (LLMs) are revolutionizing bioinformatics, enabling advanced analysis of DNA, RNA, proteins, and single-cell data. This survey provides a systematic review of recent advancements, focusing on genomic sequence modeling, RNA structure prediction, protein function inference, and single-cell transcriptomics. Meanwhile, we also discuss several key challenges, including data scarcity, computational complexity, and cross-omics integration, and explore future directions such as multimodal learning, hybrid AI models, and clinical applications. By offering a comprehensive perspective, this paper underscores the transformative potential of LLMs in driving innovations in bioinformatics and precision medicine. 
\end{abstract}

\section{Introduction}

Bioinformatics is an interdisciplinary field that combines biology, computer science, and information technology to analyze and interpret complex biological data~\cite{lu2020integrated, Abdi2024, li2024progress, jiang2025biological}. Recently, Large Language Models (LLMs) have demonstrated remarkable progress in the domain of natural language processing (NLP), with applications that span a wide array of tasks~\cite{min2023recent, jiangetal2023cognitive, raiaan2024review, jiang2025benchmarking}. However, the nature of biological data and the associated tasks differ significantly from text data, presenting unique challenges. The accurate and precise handling of biomedical data to effectively form features and embeddings suitable for LLMs is an ongoing challenge that necessitates innovative solutions~\cite{chen2022interpretable, wang2023optimized, esm2024cambrian, liu2025generalist}.

\begin{figure*}[t]
    \centering
    \includegraphics[width=\linewidth]{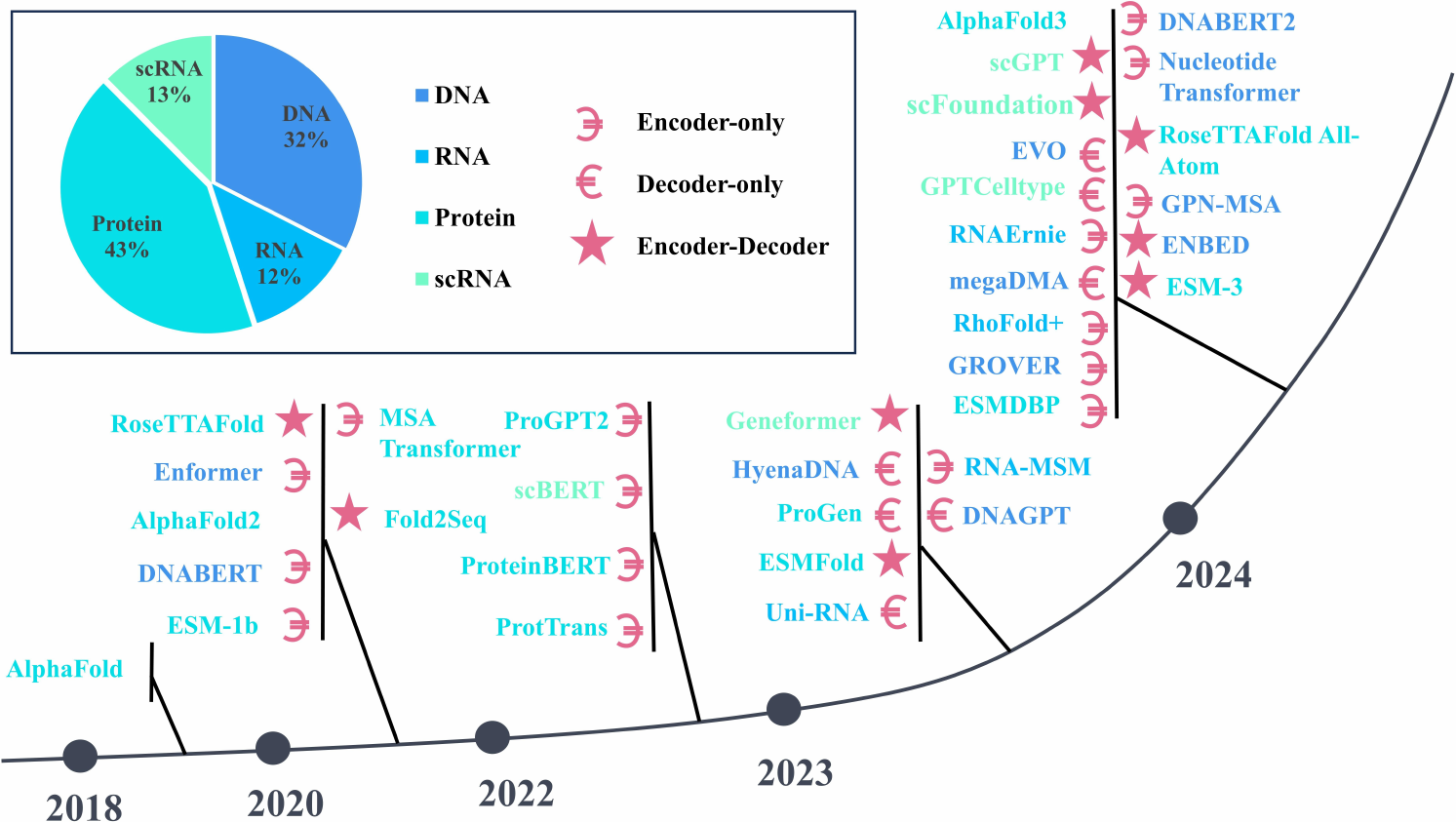}
    \caption{Selected milestones in the integration of LLMs into bioinformatics: Advances in applications to DNA, RNA, protein, and single-cell RNA (scRNA).}
    \label{Figure1_overview}
\end{figure*}

Within the biological domain, tasks exhibit a high degree of variability and specificity. These include the functional prediction and generation of DNA sequences~\cite{nguyen2024sequence}, the prediction of RNA structure and function~\cite{chen2022interpretable, shen2024accurate}, the prediction and design of protein structures~\cite{jumper2020alphafold, jumper2021highly, team2024protenix}, and the analysis of single-cell data~\cite{wolf2018scanpy}, which encompasses dimensionality reduction, clustering, cell annotation, and developmental trajectory analysis. There is a growing interest among researchers in harnessing the power of LLMs for bioinformatics and computational biology, yielding significant results~\cite{ferruz2022protgpt2, nijkamp2023progen2, li2025ds}. As illustrated in Figure~\ref{Figure1_overview}, the development, training, and application of large models in bioinformatics are increasing at a rapid pace. Despite this, the diverse methodologies focused on these various tasks have not been systematically summarized and analyzed, presenting an opportunity for comprehensive review and synthesis.

\textbf{Organization of This Survey}: This paper provides a comprehensive review of recent advancements in LLMs for bioinformatics. We begin with preliminary concepts (§2), covering key model architectures and their relevance to biological data. Next, an overview of representative LLMs applied in bioinformatics is presented in Table~\ref{tab:model_overview_1}, followed by Table~\ref{tab:llm_training_costs}, which showcases a quantitative profile of computational costs associated with training these LLMs. Table~\ref{tab:llm_summary} further synthesizes statistics derived from the surveyed models, highlighting average training duration and graphical memory usage across different model types. Subsequent sections explore LLM-driven innovations across various subdomains, DNA and Genomics (§3), RNA (§4), Proteins (§5), and Single-Cell Analysis (§6). Finally, we discuss key challenges (§7.1) and propose future directions (§7.2), emphasizing multimodal learning, hybrid AI models, and clinical applications. To conclude, we analyze the limitations of this survey, highlighting areas that require further exploration to fully capture the evolving landscape of LLMs in bioinformatics.

\section{Preliminaries}

LLMs have demonstrated remarkable advancements across various AI applications, including bioinformatics, where they enable sophisticated sequence modeling, structure prediction, and functional annotation~\cite{li2024progress}. Mechanistically, the initial design of each LLM typically follows three main architectural paradigms: encoder-only, decoder-only and encoder-decoder models. Each of these architectures has distinct advantages and is suited for different types of bioinformatics tasks. Therefore, in this section, we provide a concise overview of these architectures and corresponding relevance to their applications in bioinformatics.

\subsection{Encoder-only}

Encoder-only models, such as BERT-based architectures (e.g., ProteinBERT~\cite{brandes2022proteinbert}), primarily focus on representation learning by capturing contextual dependencies within input sequences. These models utilize bidirectional self-attention, allowing them to learn rich, contextualized embeddings which are crucial for those downstream tasks such as sequence classification, gene expression prediction, and regulatory element identification. However, encoder-only models have limitations in generative tasks, as they lack autoregressive decoding mechanisms. 

\subsection{Decoder-only}

Decoder-only models, represented by GPT-based architectures (e.g., ProGen2~\cite{nijkamp2023progen2}, Evo~\cite{nguyen2024sequence}),  operate in a casual, autogressive manner, which means that they always generate outputs token by token based on previously generated information. These models are particularly well-suited for sequence generation, structure prediction, and functional annotation, making them highly valuable in bioinformatics applications that require de novo sequence synthesis and predictive modeling. Regarding the disadvantages of decoder-only models, for example, their reliance on unidirectional attention can limit their abilities to fully capture long-range bidirectional dependencies, which are essential to understand complicated physiological reactions \textit{in vivo}. Additionally, they tend to require extensive fine-tuning when applied to domain-specific tasks, as pre-trained general-purpose models may lack sufficient knowledge of biological sequences. 

\subsection{Encoder-Decoder}

Encoder-decoder models, such as T5-based and transformer-based architectures (e.g., RoseTTAFold~\cite{baek2021accurate}), are designed for sequence-to-sequence tasks, where an input sequence is transformed into an output sequence. This architecture is particularly useful for the tasks involving mapping between different biological modalities, (e.g., gene expression predication, multiomics data integration). For RoseTTAFold, it employs a three-track neural network to predict protein interactions as well as complex formations. Meanwhile, the encoder-decoder architecture shows great potential when applied to the tasks that require bidirectional context understanding and structured output generation, represented by RNA secondary structure prediction (e.g., RhoFold+~\cite{shen2024accurate}) and genome-wide variant effect prediction. However, these models often require substantial computational resources for both training and inference, making them less accessible for researchers with limited computational infrastructure (Table~\ref{tab:llm_training_costs} and Table~\ref{tab:llm_summary}). Additionally, their performance is highly dependent on large-scale domain-specific pre-training, necessitating extensive datasets for generalization.

\newcolumntype{Y}{>{\raggedright\arraybackslash}X} 
\newcolumntype{Z}{>{\centering\arraybackslash}X}   

\begin{table*}[h!]
    \centering
    \scriptsize
    \setlength{\tabcolsep}{4pt} 
    \renewcommand{\arraystretch}{1} 
    \resizebox{\textwidth}{!}{%
        \begin{tabularx}{\textwidth}{@{}p{1.5cm}|p{2cm}|p{2cm}|p{2cm}|p{1.5cm}|p{1cm}|Z@{}}
            \hline
            \textbf{Model} & \textbf{Author\_Time} & \textbf{Venue} & \textbf{Type} & \textbf{Datasets} & \textbf{Task} & \textbf{Focus} \\
            \hline
            DNABERT & \citet{ji2021dnabert} & Bioinformatics & Encoder-only & ENCODE & DNA & Predicts genomic functions using DNA sequences \\
            \hline
            AlphaFold2 & \citet{jumper2021highly} & Nature & \textbf{---} & PDB, BFD, etc & Protein & Predicts protein 3D structures from sequences \\
            \hline
            RoseTTAFold & \citet{baek2021accurate} & Science & Encoder-Decoder & PDB, etc & Protein & Predicts protein structures and interactions efficiently \\
            \hline
            Enformer & \citet{avsec2021effective} & Nat. Methods & Encoder-only & UCSC Genome Browser, etc & DNA & Predicts gene expression from DNA sequences \\
            \hline
            ESM-1b & \citet{rives2021biological} & PNAS & Encoder-only & UniParc & Protein & Predicts protein structure and function \\
            \hline
            MSA Transformer & \citet{pmlr-v139-rao21a} & PMLR & Encoder-only & UniRef50 & Protein & Predicts protein structures using multiple sequence alignments \\
            \hline
            Fold2Seq & \citet{pmlr-v139-cao21a} & PMLR & Encoder-Decoder & CATH 4.2 & Protein & Designs protein sequences from 3D folds \\
            \hline
            ProGPT2 & \citet{ferruz2022protgpt2} & Nat. Commun. & Decoder-only & Uniref50 & Protein & Generates novel protein sequences with natural-like properties \\
            \hline
            scBERT & \citet{yang2022scbert} & Nat. Mach. Intell. & Encoder-only & Panglao & scRNA & Annotates cell types from single-cell RNA-seq data \\
            \hline
            ProteinBERT & \citet{brandes2022proteinbert} & Bioinformatics & Encoder-only & UniProtKB and UniRef90 & Protein & Predicts protein functions from sequences \\
            \hline
            ProtTrans & \citet{elnaggar2021prottrans} & IEEE PAMI & Encoder-only & UniRef and BFD & Protein & Predicts protein functions and structures \\
            \hline
            ProGen & \citet{madani2023large} & Nat. Biotechnol. & Decoder-only & UniprotKB, etc & Protein & Generates functional protein sequences across families \\
            \hline
            ESMFold & \citet{lin2023evolutionary} & Science & Encoder-Decoder & UniRef50 & Protein & Predicts protein structures from single sequences \\
            \hline
            Geneformer & \citet{theodoris2023transfer} & Nature & Encoder-Decoder & GEO, SRA, HCA, etc & scRNA & Predicts gene networks from single-cell data \\
            \hline
            HyenaDNA & \citet{nguyen2023hyenadna} & NeurIPS & Decoder-only & Human reference genome & DNA & Models long-range DNA interactions at single-nucleotide resolution \\
            \hline
            Uni-RNA & \citet{wang2023uni} & bioRxiv & Encoder-only & RNAcentral, etc & RNA & Predicts RNA structures, functions, and properties \\
            \hline
            ProGen2 & \citet{nijkamp2023progen2} & Cell Syst. & Decoder-only & UniProtKB and BFD & Protein & Generates functional protein sequences across families \\
            \hline
            xTrimoPGLM & \citet{chen2025xtrimopglm} & Nat. Methods & Encoder-Decoder & Uniref90 and ColabFoldDB & Protein & Predicts and designs protein sequences and structures \\
            \hline
            RNA-MSM & \citet{zhang2024multiple} & Nucleic Acids Res. & Encoder-only & RNAcmap & RNA & Predicts RNA structures using evolutionary information \\
            \hline
            TFBert & \citet{luo2023improving} & Interdiscip. Sci. Comput. Life Sci. & Encoder-only & 690 ChIP-seq & DNA & Predicts transcription factor binding sites \\
            \hline
            DNAGPT & \citet{zhang2023dnagpt} & arXiv & Decoder-only & GRCh38 & DNA & Generates and analyzes DNA sequences \\
            \hline
            GROVER & \citet{sanabria2024dna} & Nat. Mach. Intell. & Encoder-only & GRCh37 & DNA & Predicts DNA functions from sequence context \\
            \hline
            megaDNA & \citet{shao2024long} & Nat. Commun. & Decoder-only & NCBI GenBank & DNA & Generates and analyzes functional genomes \\
            \hline
            Nucleotide Transformer & \citet{dalla2024nucleotide} & Nat. Methods & Encoder-only & GRCh38 & DNA & Predicts molecular phenotypes from DNA sequences \\
            \hline
            RNAErnie & \citet{wang2024multi} & Nat. Mach. Intell. & Encoder-only & RNAcentral & RNA & Predicts RNA functions and structures \\
            \hline
            RhoFold+ & \citet{shen2024accurate} & Nat. Methods & Encoder-only & RNAcentral & RNA & Predicts RNA 3D structures from sequences \\
            \hline
            GPTCelltype & \citet{hou2024assessing} & Nat. Methods & Decoder-only & figshare, Zenodo, GEO, etc. & scRNA & Automates cell type annotation using GPT-4 \\
            \hline
            Evo & \citet{nguyen2024sequence} & Science & Decoder-only & OpenGenome & DNA & Predicts and designs DNA, RNA, proteins \\
            \hline
            scFoundation & \citet{hao2024large} & Nat. Methods & Encoder-Decoder & HCA, Single Cell Portal, GEO, etc & scRNA & Predicts and analyzes single-cell transcriptomics data \\
            \hline
            scGPT & \citet{cui2024scgpt} & Nat. Methods & Encoder-Decoder & CELLxGENE & scRNA & Predicts and analyzes single-cell omics data \\
            \hline
            AlphaFold3 & \citet{abramson2024accurate} & Nature & \textbf{---} & PDB & Protein & Predicts biomolecular structures and interactions accurately \\
            \hline
            DNABERT2 & \citet{zhou2023dnabert} & ICLR & Encoder-only & Human and multi-species genome & DNA & Predicts genomic functions across species efficiently \\
            \hline
            ESM-DBP & \citet{zeng2024improving} & Nat. Commun. & Encoder-only & UniProtKB & Protein & Predicts DNA-binding proteins and residues accurately \\
            \hline
            RoseTTAFold All-Atom & \citet{krishna2024generalized} & Science & Encoder-Decoder & PDB, etc & Protein & Predicts and designs biomolecular structures \\
            \hline
            ProstT5 & \citet{heinzinger2024bilingual} & NAR Genom. Bioinform. & Encoder-Decoder & AFDB & Protein & Translates protein sequences to 3D structures \\
            \hline
            EpiGePT & \citet{gao2024epigept} & Genome Biol. & Encoder-only & ENCODE & DNA & Predicts context-specific epigenomic signals and interactions \\
            \hline
            RiNALMo & \citet{penic2024rinalmo} & arXiv & Encoder-only & RNAcentral & RNA & Predicts RNA structures and functions \\
            \hline
            ENBED & \citet{malusare2024understanding} & Bioinform. adv. & Encoder-Decoder & NCBI-Genome & DNA & Analyzes DNA sequences with byte-level precision \\
            \hline
            GPN-MSA & \citet{benegas2025dna} & Nat. Biotechnol. & Encoder-only & multiz MSA, etc & DNA & Predicts genome-wide variant effects efficiently \\
            \hline
            ESM-3 & \citet{hayes2025simulating} & Science & Encoder-Decoder & UniRef & Protein & Predicts and designs proteins with multi-modal inputs \\
            \hline
        \end{tabularx}
    }
    \caption{Comprehensive overview of representative LLMs in bioinformatics, systematically categorized by model architecture, dataset, task, and application domain. This table summarizes key advances in LLM-driven methods across major bioinformatics modalities, including DNA, RNA, protein, and single-cell omics. Dashes ("\textbf{---}") indicate information not applicable for the respective model.}
    \label{tab:model_overview_1}
\end{table*}

\section{DNA and Genomics: Learn and Generate }
The research landscape of LLMs in genomics is witnessing a surge in development, particularly in their application across a spectrum of genomic tasks. These models not only improve the analytical capabilities of DNA sequence analysis, but also accurately predict the impact of genetic mutations, identify key regulatory sequences, and thus advance the understanding of genomic functions. Moreover, the generation of biologically functional gene sequences using LLMs is an area worth exploring, with potential implications for synthetic biology and gene therapy.

The integration of LLMs in genomics has opened new avenues for research, offering insights into the complex interplay between genetic sequences and their biological functions. For instance, in the prediction of gene regulatory elements, LLMs have been shown to outperform traditional machine learning algorithms, providing a more nuanced understanding of gene regulation and expression~\cite{koido2024fundamentals}. Additionally, their application in the prediction of the effects of genetic mutations has led to a better comprehension of disease mechanisms and potential therapeutic targets.

DNABERT is a pre-trained bidirectional encoder representation, which can capture global and transferrable understanding of genomic DNA sequences based on up and downstream nucleotide contexts. It can be fined tuned to many other sequence analyses tasks such as predict proximal and core promoter regions, identify transcription factor binding sites, recognize canonical and non-canonical splice sites and identify functional genetic variants~\cite{ji2021dnabert}.

DNABERT-2 is a high-performance foundation model based on the Transformer architecture, designed for multi-species genome analysis~\cite{zhou2023dnabert}. By incorporating innovative tokenization methods, efficient attention mechanisms, and a comprehensive evaluation benchmark, it provides a powerful tool for understanding and interpreting genomic data across diverse species~\cite{feng2024llmbased}.

GeneBERT is a large-scale pre-trained model that integrates diverse genomic data modalities across multiple cell types in a self-supervised manner. By leveraging multi-modal genome data during pre-training, GeneBERT captures complex biological patterns and relationships, enabling robust performance on a wide range of downstream tasks. These tasks include promoter prediction, transcription factor binding site prediction, disease risk estimation, and RNA splicing analysis. Its ability to generalize across various genomic contexts and tasks makes GeneBERT a powerful tool for advancing genome research and precision medicine~\cite{mo2021multi}.

GROVER is a deep learning model designed to capture the structural and contextual features of DNA sequences by simultaneously learning token-level characteristics and broader sequence contexts. It demonstrates superior performance in both next-k-mer prediction and fine-tuning tasks, such as promoter identification and DNA-protein binding prediction. GROVER's ability to effectively model DNA language structure makes it a valuable tool for advancing genomic research and understanding regulatory mechanisms~\cite{sanabria2024dna}.

MegaDNA is a groundbreaking long-context generative model designed for genomic sequence analysis and de novo sequence generation. This model leverages a multi-scale Transformer architecture to process and generate DNA sequences at single-nucleotide resolution, enabling unprecedented capabilities in genomics research~\cite{shao2024long}.

In the realm of gene sequence generation, LLMs hold promise for the creation of sequences with specific biological functions. This capability could facilitate the design of genes for targeted therapies, the enhancement of crop traits, and the development of biomanufacturing processes. The potential of LLMs in this domain is vast, and ongoing research is expected to unveil novel applications and techniques.

Nucleotide Transformer is a state-of-the-art foundation model designed for genomic sequence analysis, leveraging large-scale pre-training on diverse DNA and RNA sequences to capture complex biological patterns. Developed to address the challenges of understanding genomic language, this model integrates advanced deep learning techniques with extensive genomic datasets, enabling robust performance across a wide range of downstream tasks~\cite{dalla2024nucleotide}.

Evo is a groundbreaking genomic foundation model designed to predict and generate biological sequences—spanning DNA, RNA, and proteins—from molecular to genome scales. Developed by researchers at the Arc Institute and Stanford University, Evo leverages advanced deep learning architectures and large-scale pre-training to achieve unprecedented capabilities in biological sequence analysis and design~\cite{nguyen2024sequence}.

\section{RNA: Structure and Function}

\subsection{RNA Structure}
RNA molecules play critical roles in biological systems, functioning as catalytic ribozymes, metabolite-sensing riboswitches, and epigenetically regulatory long noncoding RNAs~\cite{zhang2022advances}. These diverse functions are enabled by the ability of single-stranded RNA to fold into intricate secondary and tertiary structures. Accurate prediction of RNA structures is therefore essential for understanding their biological mechanisms and therapeutic potential.

However, RNA structure prediction remains challenging due to the complexity and dynamic nature of RNA folding. RNA molecules exhibit conformational flexibility, long-range interactions, and non-canonical base pairing, making their 3D structures difficult to model. Additionally, the scarcity of high-quality experimental RNA structure data limits the training and development of LLMs for this purpose.

Recent advances in computational methods, including deep learning and physics-based simulations, have significantly improved RNA structure prediction. Here, we conclude some RNA-related LLMs recently to better understand developments and limitations in RNA structure predction.

\subsubsection{RNA secondary structure prediction}

Some researchers benchmarked 6 RNA-LLMs which include RNABERT, RNA-FM, RNA-MSM, ERNIE-RNA, RNAErnie and RiNALMo for their ability to predict RNA secondary structure. They found that RiNALMo and ERNIE-RNA were the models that could better represent and separate the RNA families in the projection without almost overlap~\cite{zablocki2024comprehensive}.

\subsubsection{RNA Teridry structure prediction}

Uni-RNA represents a paradigm shift in RNA research by combining large-scale pre-training with advanced deep learning techniques. Its ability to accurately predict RNA structures, functions, and properties positions it as a powerful tool for accelerating discoveries in RNA biology and therapeutic development~\cite{wang2023uni}.

RhoFold+ is an advanced method leveraging deep learning and RNA language models to efficiently and accurately predict RNA three-dimensional structures. Its core strength lies in the integration of a large-scale pre-trained RNA language model (RNA-FM) with a deep learning architecture, enabling end-to-end prediction from RNA sequences to 3D structures~\cite{shen2024accurate}.

NuFold is a state-of-the-art deep learning model designed for the accurate prediction of RNA tertiary structures. It addresses the significant gap between RNA sequence data and experimentally determined structures by leveraging advanced computational techniques~\cite{kagaya2025nufold}.

\subsection{RNA Function}

BEACON comprises 13 distinct tasks derived from extensive prior research, covering structural analysis, functional studies, and engineering applications~\cite{ren2024beacon}. Functional tasks focus on the biological roles of RNA, including splice site prediction and non-coding RNA function classification.

RNA-RNA interactions are involved in post-transcriptional processes, contributing to gene expression regulation~\cite{wang2022emerging}. RNA-protein interactions are vital for maintaining cellular homeostasis~\cite{gallardo2024regulatory}, and disruptions in these interactions can lead to cellular dysfunctions or diseases such as cancer~\cite{liu2020classification}. Furthermore, RNA-small molecule interactions have significant implications in therapeutic development, as RNAs can serve as potential drug targets, especially when conventional protein targets are less accessible~\cite{kaur2024rna}.

BioLLMNet provides a way to transform feature space of different molecule’s language model features and uses learnable gating mechanism to effectively fuse features and achieves state-of-the-art performance in RNA-protein, RNA-small molecule, and RNA-RNA interactions, outperforming existing methods in RNA-associated interaction prediction~\cite{tahmid2024biollmnet}.

\subsection{RNA sequence generation}

RNA-GPT, a multi-modal RNA chat model designed to simplify RNA discovery by leveraging extensive RNA literature. RNA-GPT integrates RNA sequence encoders with linear projection layers and state-of-the-art LLMs for precise representation alignment, enabling it to process user-uploaded RNA sequences and deliver concise, accurate responses~\cite{xiao2024rna}.

RNA-DCGen, a generalized framework for RNA sequence generation that is adaptable to any structural or functional properties through straightforward fine-tuning with an RNA language model (RNA-LM).To address these challenges: specialization for fixed constraint types, such as tertiary structures, and lack of flexibility in imposing additional conditions beyond the primary property of interest.~\cite{shahgir2024rna}

\section{Protein: Prediction and Design}
Recently, LLMs have emerged as transformative computational tools in protein research, demonstrating remarkable potential to advance both fundamental comprehension and applied engineering of protein structures and functions. In this section, we introduce representative protein-related LLMs, as categorized in Table~\ref{tab:model_overview_1}, based on their model architectures and diverse protein research applications, such as protein structure and function prediction and protein generation and design.

\subsection{Protein Structure and Function}
AlphaFold2 employs deep learning to predict protein 3D structures with atomic-level accuracy, achieving unprecedented success in CASP14. Its open-source database has revolutionized structural biology, enabling rapid drug discovery and mechanistic studies across biomedical research~\cite{jumper2021highly}.

RoseTTAFold integrates sequence, distance, and 3D coordinate predictions through a three-track neural architecture. It also achieves near-experimental accuracy in CASP14, enabling rapid modeling of understudied proteins for therapeutic and evolutionary analyses~\cite{baek2021accurate}.

ESM-1b leverages a transformer-based encoder architecture to infer protein tertiary structures and functional characteristics through self-supervised learning on large-scale protein sequence databases, without relying on manual annotations of the sequence~\cite{rives2021biological,meier2021language}.

ProteinBERT separates local (character-level) and global (sequence-level) representations and advances transformer architecture through a self-supervised learning paradigm by establishing a unified framework for multitask protein analysis~\cite{brandes2022proteinbert}.

ProtTrans, a transformer-based protein language model, employs self-supervised learning on 100+ million sequences to capture evolutionary-structural patterns. It achieves state-of-the-art performance in tertiary structure prediction, functional annotation, and engineering design while enabling efficient transfer learning across diverse proteomic tasks~\cite{elnaggar2021prottrans}.

AlphaFold3 advances structural biology by integrating geometric deep learning with diffusion models, achieving atomic-resolution predictions for generalized biomolecular complexes (proteins, DNA, ligands). It demonstrates better accuracy in ligand binding sites over experimental methods, revolutionizing drug discovery and systems biology through whole-cell interactome modeling~\cite{abramson2024accurate}.

ESM-DBP integrates protein language models with DNA-binding specificity prediction, leveraging evolutionary-scale sequence training to accurately identify DNA-interaction motifs and binding sites~\cite{zeng2024improving}.

RoseTTAFold All-Atom is a fast and accurate neural network for predicting diverse biomolecular assemblies. It models proteins, nucleic acids, small molecules, metals, and covalent modifications, advancing structural biology and drug discovery~\cite{krishna2024generalized}.

\subsection{Protein Design and Engineering}

LLMs will be well suited for protein design applications, such as designing antibodies with reduced aggregation propensity, development of drugs that target specific protein phases in diseases, and understanding the mechanisms behind diseases caused by protein misfolding and aggregation.

ProtGPT2 is a pretrained transformer-based language model for protein sequence generation and engineering. It explores new protein regions while preserving natural protein features, offering high-throughput design capabilities~\cite{ferruz2022protgpt2}.

ProGen2 is a protein language models with 6.4B parameters, trained on over a billion proteins. It excels in capturing sequence distribution, generating novel sequences, and predicting protein fitness without fine-tuning~\cite{nijkamp2023progen2}.

ESM-3 is a large language model trained on protein sequences, structures, and functions. It offers multimodal analysis, generating novel proteins and predicting 3D structures, advancing drug discovery and biotechnology~\cite{hayes2025simulating}.

\section{scRNA: Development and Challenge}
Single-cell sequencing technology, an evolution of transcriptome sequencing, enables the examination of gene expression and transcription levels at the individual cell level~\cite{potter2018single}. This technology is pivotal for understanding various biological processes at the cellular level, such as disease progression, therapeutic efficacy, and resistance. It can identify relevant cell subpopulations and molecules, holding profound significance for the advancement of bioinformatics. However, the effective and accurate processing and analysis of vast single-cell data sets present a challenge for bioinformaticians. Traditional analysis methods and pipelines are based on tools like Seurat and Scanpy~\cite{hao2024dictionary, wolf2018scanpy}. With the advancement of deep learning, numerous studies have integrated frameworks such as deep learning and large language models with the processing and analysis of single-cell data, achieving significant progress and demonstrating the immense potential of combining these approaches to faster development in the field of single-cell research~\cite{molho2024deep}.

scBERT is a pioneering deep learning model designed for cell type annotation in scRNA-seq data~\cite{yang2022scbert}. It adapts the BERT framework, originally developed for NLP, to the domain of single-cell transcriptomics. The model leverages a pre-training and fine-tuning paradigm, where it first learns general patterns of gene-gene interactions from large-scale unlabeled scRNA-seq data and then fine-tunes on specific tasks using labeled data to predict cell types~\cite{molho2024deep}.

Geneformer is a foundational deep learning model based on the Transformer architecture, specifically designed for analyzing scRNA-seq data. Pre-trained in a massive corpus of 29.9 million single-cell human transcriptomes, Geneformer uses self-supervised learning to capture gene-gene interactions and regulatory dynamics in various cell types and tissues~\cite{theodoris2023transfer}.

GPTCelltype is an innovative R software package designed to automate cell type annotation in scRNA-seq analysis by leveraging the capabilities of GPT-4, a state-of-the-art LLMs. This tool represents a significant advancement in the field of bioinformatics, offering a cost-effective and efficient alternative to traditional manual or semi-automated annotation methods~\cite{hou2024assessing}.

scFoundation is a groundbreaking large-scale pre-trained model designed for scRNA-seq data analysis, featuring 100 million parameters and capable of handling approximately 20,000 genes simultaneously. Pre-trained in more than 50 million human single-cell transcriptomic profiles, scFoundation represents a significant advancement in the field of single-cell genomics, offering a robust framework for diverse downstream tasks such as gene expression enhancement, drug response prediction, and cell type annotation~\cite{hao2024large}.

scGPT is a state-of-the-art foundation model designed for single-cell multi-omics data analysis, leveraging the transformer architecture to capture complex gene-cell interactions. Pre-trained in over 33 million human single-cell transcriptomes from the CELLxGENE database, scGPT excels in extracting meaningful biological insights and generalizing across diverse downstream tasks, such as cell type annotation, perturbation prediction, batch integration, and gene regulatory network inference~\cite{cui2024scgpt}.

Foundation models like scBERT~\cite{yang2022scbert}, and scGPT~\cite{cui2024scgpt} have transformed single-cell data analysis through self-supervised pretraining on millions of transcriptomes, enabling robust transfer learning for tasks such as cell annotation, perturbation prediction, and batch correction. Architectural innovations (e.g., performer encoders, rank-based encoding) address computational challenges of high-dimensional data, while task-specific designs improve generalizability and interpretability via attention mechanisms.

However, limitations persist, including trade-offs between computational efficiency and expression resolution, biases from imbalanced pretraining data, and the "black-box" nature of Transformer architectures. High computational costs further hinder accessibility. Future efforts should prioritize multimodal integration of transcriptomic, epigenetic, and spatial data, knowledge-guided architectures incorporating biological networks, efficient few-shot learning frameworks, and lightweight adaptations for greater usability. Addressing these challenges will enhance biological interpretability and bridge AI capabilities with actionable insights, advancing precision medicine and systems biology.

\section{Conclusions and Future Directions}

In conclusion, this paper comprehensively examined the applications of LLMs across DNA, RNA, protein, and single-cell data analysis, highlighting key research contributions and emerging methodologies. Despite advances, LLM applications in bioinformatics remain evolving, requiring key challenges to address for full potential. Therefore, we here discuss the current limitations and outline promising future directions for advancing LLM-driven bioinformatics research. 

\subsection{Challenges and Limitations}

\subsubsection{Data Scarcity, Quality and Bias}

LLMs always require large-scale, high-quality biological datasets for effective training, yet annotated genomic, transcriptomic, and proteomic data remain limited~\cite{lu2024large}. Unlike natural language corpora, which are abundant and diverse, biological datasets are often noisy, incomplete, or biased toward well-studied species and diseases. Consequently, model generalizability suffers, leading to biased predictions which may not hold across diverse biological contexts. Additionally, batch effects and experimental noise complicate the development of robust foundation models for bioinformatics~\cite{yu2024assessing}.

\subsubsection{Computational Complexity and Model Efficiency}

State-of-the-art LLMs, such as AlphaFold and DNABERT, require massive computational resources for both training and inference. This computational barrier limits accessibility, particularly for research groups with limited infrastructure (Table \ref{tab:llm_summary} and Table \ref{tab:llm_training_costs}). Furthermore, relatively longer biological sequences significantly increase memory and processing requirements, making it challenging to apply standard Transformer architectures to genome-scale data~\cite{bernard2025rna}. Efficient methods of model compression and retrieval-augmented need to be further explored to enhance scalability.

\subsubsection{Multimodal Learning and Cross-Omics Integration }

Biological systems exhibit intricate interactions across multiple molecular layers, including but not limited to genomics and metabolomics. Despite recent advancements, current LLMs remain predominantly trained on single-modality datasets, constraining their ability to model cross-scale molecular dependencies. Addressing this limitation requires the development of multimodal architectures capable of integrating heterogeneous biological data in a biologically meaningful and computationally efficient manner~\cite{dankan2025future}.

\subsection{Future Directions}

To overcome the challenges mentioned above, future research should focus on developing efficient, interpretable, and multimodal LLM architectures tailored for bioinformatics. 

\subsubsection{Hybrid AI Models for Biological Reasoning }

Integrating LLMs with mechanistic models, represented by graph neural networks (GNNs), and knowledge graphs could improve biological reasoning and interpretability~\cite{feng2025knowledge}. Hybrid approaches that combine deep learning with symbolic AI~\cite{colelough2025neuro} or constraint-based modeling\cite{bystrova2024causal} may enable causality-aware predictions in biological systems.

\subsubsection{Multimodal and Cross-Omics Integration}

Future LLMs should be designed with multimodal learning capabilities, enabling the simultaneous processing of DNA, RNA, protein, and epigenetic data. By integrating self-supervised learning across diverse omics datasets, these models could enhance biological understanding and improve cross-species generalization. Furthermore, to enhance explainability, biologically constrained LLMs incorporating evolutionary principles and regulatory networks could improve model robustness and reliability~\cite{feng2023gene}, ensuring greater alignment with fundamental biological mechanisms.

\subsubsection{Towards Clinical and Biomedical Applications}

Bridging the gap between AI-driven bioinformatics and real-world biomedical applications necessitates further advancements in model validation, regulatory compliance, and ethical considerations. Moving forward, LLM tools require rigorous clinical evaluation and experimental benchmarking to ensure healthcare reliability and safety~\cite{perlis2023evaluating}.

In summary, addressing these challenges and research directions will enable next-generation LLMs to drive transformative breakthroughs in genomics and precision medicine, paving the way for a new era of AI-driven biological discovery.

\begin{table}[t]
\centering
\scriptsize
\setlength{\tabcolsep}{4pt}
\renewcommand{\arraystretch}{1}
\resizebox{0.5\textwidth}{!}{%
\begin{tabular}{@{}p{3cm}|p{2.5cm}|p{2.5cm}@{}}
\hline
\textbf{Model Type} & \textbf{Avg. Graphical Memory per Device}& \textbf{Avg. Training Duration}\\
\hline
Encoder-only     & \textasciitilde43 GB& \textasciitilde14 Days\\
\hline
Decoder-only     & \textasciitilde46 GB& \textasciitilde5 Days\\
\hline
Encoder–Decoder  & \textasciitilde81 GB& \textasciitilde40 Days\\
\hline
\end{tabular}
}
\vspace{0.5em}
\caption{Statistics of average training duration and graphical memory per device across different LLM architectures. Values are computed based on surveyed models.}
\label{tab:llm_summary}
\end{table}

\section*{Limitations}

In this paper, we provide a survey of LLMs in bioinformatics. Despite our best efforts, there may be still several limitations that remain in this study.

\textbf{Scope Restriction:} Our survey focuses on four subdomains: DNA, RNA, protein, and single-cell analysis. However, other areas, represented by epigenomics and metagenomics,  are not covered in depth. Expanding the scope to include these fields would provide a more holistic perspective on LLM-driven bioinformatics research. 

\textbf{Rapidly Evolving Field:} Given the rapid advancements in LLMs, some recent breakthroughs may not be fully captured in this survey. The field is continuously evolving, with new models and methodologies emerging at a fast pace. 

\textbf{Lack of Empirical Benchmarking:} While this paper synthesizes and analyzes existing research, it does not include direct experimental validation or standardized benchmarking of LLMs for bioinformatics tasks. Specifically, a rigorous assessment of LLMs' performance under consistent conditions, such as dataset standardization and computational efficiency, remains an open challenge.

Moving forward, we will actively monitor and incorporate emerging developments to ensure that future iterations of this work reflect the most up-to-date progress in LLMs applications for bioinformatics.

\section*{Ethical Statement}

There are no ethical issues.

\section*{Acknowledgements}

We want to thank our anonymous AC and reviewers for their feedback. This study was supported by the Chinese University of Hong Kong (CUHK; award numbers 4937025, 4937026, 5501517 and 5501329 to Y.L.) and the IdeaBooster Fund (IDBF23ENG05 and IDBF24ENG06 to Y.L.) and partially supported by a grant from the Research Grants Council of the Hong Kong Special Administrative Region (Hong Kong SAR), China (project no. CUHK 24204023 to Y.L.) and a grant from the Innovation and Technology Commission of the Hong Kong SAR, China (project no. GHP/065/21SZ and ITS/247/23FP to Y.L.). This research was also funded by the Research Matching Grant Scheme at CUHK (award numbers 8601603 and 8601663 to Y.L.) from the Research Grants Council, Hong Kong SAR, China.

\bibliography{custom}

\appendix
\section{Appendix}

\begin{table*}[!t]
\centering
\scriptsize
\setlength{\tabcolsep}{4pt}
\renewcommand{\arraystretch}{1}
\begin{tabular}{@{}p{3.2cm}|p{4cm}|p{3.5cm}|p{2cm}@{}}
\hline
\textbf{Model} & \textbf{Computational Resources} & \textbf{Graphical Memory per Device} & \textbf{Training Duration} \\
\hline
DNABERT         & 8 * NVIDIA 2080Ti& 11 GB& \textbf{---}\\
\hline
AlphaFold2      & Tensor Processing Unit (TPU) v3 & 32 GiB& \textasciitilde10 Days\\
\hline
RoseTTAFold     & NVIDIA V100& 32 GB& \textasciitilde28 Days\\
\hline
Enformer        & 64 * TPU v3& 32 GiB& \textasciitilde3 Days\\
\hline
ESM-1b          & \textbf{---}& \textbf{---}& \textbf{---}\\
\hline
MSA Transformer & 32 * NVIDIA V100& 32 GB& \textbf{---}\\
\hline
Fold2Seq        & 2 * Tesla K80& 24 GB& \textbf{---}\\
\hline
ProGPT2         & 128 * NVIDIA A100&                   80 GB&  \textasciitilde4 Days\\
\hline
scBERT          & \textbf{---}& \textbf{---}& \textbf{---}\\
\hline
ProteinBERT     & Nvidia Quadro RTX 5000& 16 GB& 28 Days\\
\hline
ProtTrans       & NVIDIA V100& 16 GB& \textbf{---}\\
\hline
ProGen          & TPU v3& 32 GiB& \textasciitilde14 Days\\
\hline
ESMFold         & NVIDIA V100& 32 GB& \textbf{---}\\
\hline
Geneformer      & 12 * NVIDIA V100& 32 GB& \textasciitilde3 Days\\
\hline
HyenaDNA        & 1 * NVIDIA A100& 40 GB& \textasciitilde80 Minutes\\
\hline
Uni-RNA         & 128 * NVIDIA A100& 80 GB& \textbf{---}\\
\hline
ProGen2         & TPU v3& 32 GiB& \textbf{---}\\
\hline
 xTrimoPGLM & 96 * DGX-A100& 320 GB&\textasciitilde150 Days\\
\hline 
 RNA-MSM & 8 * NVIDIA V100& 32 GB&\textbf{---}\\
 \hline
 TFBert & 1 * NVIDIA V100& 32 GB& \textasciitilde14 Days\\
 \hline
 DNAGPT & 8 * NVIDIA V100& 32 GB& \textasciitilde1 Day\\
 \hline
 GROVER & NVIDIA A100& 80 GB& \textbf{---}\\
 \hline
 megaDNA & NVIDIA A100 + 3090Ti& 40 GB (A100) + 24 GB (3090Ti)& \textbf{---}\\
 \hline
 Nucleotide Transformer & 128 * NVIDIA A100& 80 GB& \textasciitilde28 Days\\
 \hline
 RNAErnie & 4 * NVIDIA V100& 32 GB& \textasciitilde10 Days\\
 \hline
 RhoFold+ & 8 * NVIDIA A100& 80 GB& \textasciitilde30 Days\\
 \hline
 GPTCelltype & \textbf{---}& \textbf{---}& \textbf{---}\\
 \hline
 Evo & 64 * NVIDIA H100 + 128 * NVIDIA A100& 80 GB (H100 + A100)& \textbf{---}\\
 \hline
 scFoundation & NVIDIA Ampere GPU& \textbf{---}& \textbf{---}\\
 \hline
 scGPT & NVIDIA A100& 80 GB& \textbf{---}\\
 \hline
 AlphaFold3 & NVIDIA A100& 80 GB& \textbf{---}\\
 \hline
 DNABERT2 & 8 * NVIDIA 2080Ti& 11 GB& \textasciitilde14 Days\\
 \hline
 ESM-DBP & 4 * NVIDIA V100& 16 GB& \textasciitilde3 Days\\
 \hline
 RoseTTAFold All-Atom & 8 * NVIDIA A6000& 48 GB& \textasciitilde8 Days\\
 \hline
 ProstT5 & 8 * NVIDIA A100& 80 GB& \textasciitilde10 Days\\
 \hline
 EpiGePT & NVIDIA 3090 + NVIDIA 4090& 24 GB (3090 + 4090)& \textbf{---}\\
 \hline
 RiNALMo & 7 * NVIDIA A100& 80 GB& \textasciitilde14 Days\\
 \hline
 ENBED & 8 * NVIDIA A100& 80 GB& \textasciitilde480 GPU Hours\\
 \hline
 GPN-MSA & 4 * NVIDIA A100& 80 GB& \textasciitilde3.5 Hours\\
 \hline
 ESM-3 & \textbf{---}& \textbf{---}& \textbf{---}\\
 \hline
\end{tabular}
\vspace{0.5em}
\caption{Quantitative overview of computational costs for selected LLMs applied to bioinformatics. ("\textbf{---}" denotes unreported data. 1 GiB = 1.07 GB)}
\label{tab:llm_training_costs}
\end{table*}

\end{document}